\documentclass[11pt,letterpaper]{article}
\usepackage[T1]{fontenc}
\usepackage[utf8]{inputenc}
\usepackage[margin=1in]{geometry}
\usepackage{mathptmx,courier}
\usepackage{amsmath,amssymb,booktabs,graphicx}
\usepackage[dvipsnames]{xcolor}
\usepackage{microtype}
\usepackage[numbers,sort&compress]{natbib}
\usepackage[hyphens]{url}
\usepackage{xurl}
\usepackage{enumitem}
\usepackage[font=small,labelfont=bf,labelsep=period]{caption}
\definecolor{linkblue}{rgb}{0.10,0.27,0.46}
\usepackage[breaklinks,colorlinks,allcolors=linkblue]{hyperref}
\usepackage[capitalize]{cleveref}
\newcommand{\ind}{\mathbf{1}}
\newcommand{\doiurl}[1]{\href{https://doi.org/#1}{\nolinkurl{#1}}}
\setlist[itemize]{leftmargin=1.2em,itemsep=2pt,topsep=3pt}
\date{}
\title{Auditing Semantic and Relational Reliability in\\Open-Vocabulary Human-Object Interaction Evaluation}
\author{
S M Asif Hossain
\and
Ruksat Khan Shayoni
\and
Md Kishor Morol
}
\begin{document}
\maketitle

\begin{abstract}
Open-vocabulary human-object interaction detectors describe interactions in free text, and this creates a problem for evaluation. Category scoring rejects valid paraphrases, while semantic similarity scoring can reward predictions that name the wrong action or reverse the roles of the participants. We present a diagnostic protocol that holds every prediction fixed and scores it under five views, namely official category evaluation, an identity control, lexical normalization, graded semantic credit, and strict propositional equivalence. Comparing these views separates the effect of the matching engine from the effect of semantic credit. The protocol pairs controlled transformations that preserve or break meaning with a blinded audit of natural predictions, and we derive consistency checks that link mean credit, invalid acceptance, and transformation counts. Applied to six systems on HICO-DET, the views disagree on which system leads. GEN-VLKT leads under official evaluation, SL-HOI leads under SHOE, and GEN-VLKT leads again under strict equivalence. Semantic gains range from 1.6 to 20.3 mAP, and AgentHOI keeps less than half of its gain once equivalence is enforced. We recommend that semantic scores be reported together with relational validity and paired uncertainty, so that recovered credit is not read as better interaction understanding.
\end{abstract}

\section{Introduction}
\begin{figure}[t]
\centering
\includegraphics[width=\textwidth]{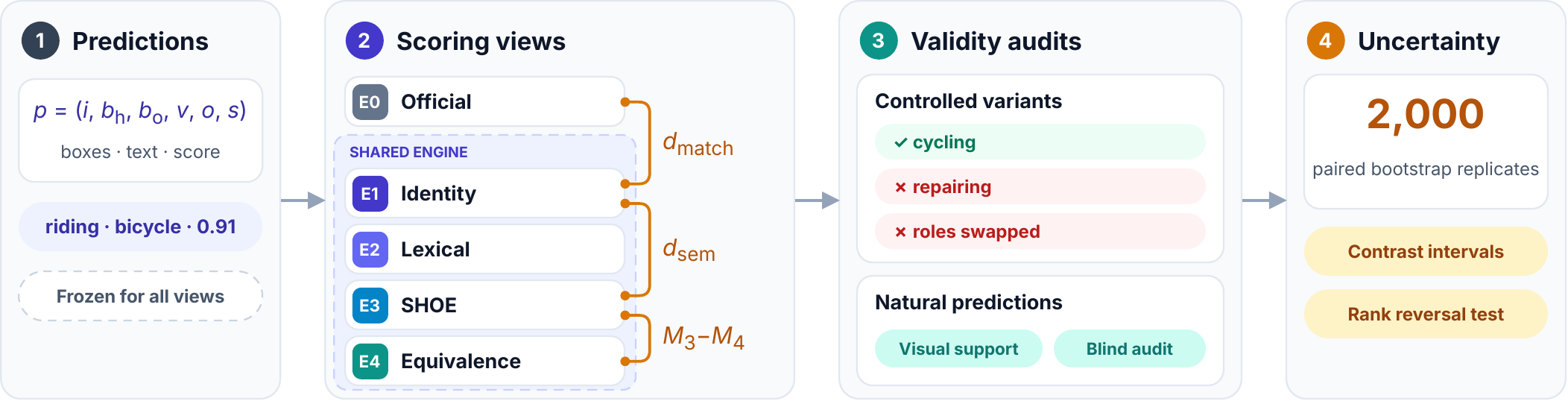}
\caption{Overview of the methodology. Predictions are frozen and scored under five views. E0 and E1 differ only in the matching engine, while E1 to E4 share one engine and differ only in how semantic credit is given. The orange brackets mark the contrasts we report. Controlled variants and a blind audit of natural predictions test whether the credit is valid, and a paired bootstrap gives intervals for every contrast.}
\label{fig:protocol}
\end{figure}

Human-object interaction detection, or HOI detection, asks a model to find a person and an object in an image and to name the action that connects them. HICO-DET has long served as the standard benchmark for this task and evaluates detections against a fixed ontology of interaction categories~\cite{chao2018}. Recent work has moved beyond such fixed ontologies. Larger vocabularies and natural-language supervision let detectors describe interactions they were never explicitly trained on~\cite{wang2021,wang2022}, and many current systems output free text rather than a category index.

This shift makes evaluation sensitive to wording, and it does so in two opposite directions. A category evaluator can reject a correct prediction simply because it is phrased differently from the annotation. A similarity-based evaluator can make the reverse mistake and accept a wrong prediction because it sounds close to the annotation. Both errors change the reported score, yet neither reflects what the model actually understood about the image.

A simple example shows the difference. If the reference is ``riding a bicycle,'' the prediction ``cycling'' describes the same interaction and should receive credit. The prediction ``repairing a bicycle'' keeps the object but changes the action, so it should not. Relations can fail in a subtler way. Reversing the participants in ``one person follows another'' keeps every word but changes who is doing what. These wrong predictions are lexically close to their references, which is exactly why a similarity score may reward them. A reliable semantic evaluator therefore has to tell these cases apart while keeping localization and entity identity fixed. There is a further complication. A prediction that matches no annotation may describe an additional valid interaction that annotators missed, and no comparison of text alone can detect this.

Recent work has started to address open-vocabulary evaluation. SHOE introduces graded semantic credit for HOI detection~\cite{noack2026}, CrossHOI-Bench curates positive and negative answer options~\cite{lei2026cross}, and diagnostic studies expose detector errors and scene-dependent failures~\cite{zhu2025,wang2026diag}. Two questions remain open. The first is how much of a score change under a new evaluator comes from semantic credit, and how much comes from the matching engine that implements it. The second is whether the extra credit corresponds to interactions that are actually valid. This paper addresses both questions with the following contributions.
\begin{itemize}
\item We define five scoring views that are applied to the same frozen predictions. Because only the evaluator changes, differences between views separate matching-engine effects from semantic credit.
\item We design complementary audits. Controlled transformations test whether an evaluator keeps credit for equivalent wording and removes it for invalid changes, and a blinded audit of natural predictions tests visual support. We also derive consistency checks for the audit records.
\item We compare six HOI systems on HICO-DET under all five views, show that their ranking depends on the evaluator, and specify the paired analyses needed to test whether such ranking changes are statistically supported.
\end{itemize}

The comparison yields three main observations. The leading system changes twice as the evaluator moves from official scoring to SHOE and then to strict equivalence. Several systems lose more than four points when moved to the shared engine with identity credit, and lexical aliases recover most of this loss, so these drops should not be read as language effects. Finally, the semantic gains differ by more than a factor of ten across systems, and the system with the largest gain loses the largest share of it under a strict reading.

\section{Related work}
\paragraph{HOI detection beyond a fixed vocabulary.}
GEN-VLKT improves the association between people and objects and transfers visual and linguistic knowledge from a pretrained vision-language model into the detector~\cite{liao2022}. EZ-HOI and HOLa adapt vision-language models for zero-shot HOI detection, the first through guided prompt learning and the second through low-rank feature adaptation~\cite{lei2024ez,lei2025hola}. THID learns a transferable detector from natural-language supervision~\cite{wang2022}, and SL-HOI streamlines the detection pipeline for open-vocabulary use~\cite{sun2026}. AgentHOI goes further and combines a multimodal large language model with grounding, without any HOI-specific training~\cite{lei2026agent}. These systems differ in their supervision, in how much of the vocabulary they see during training, and in whether they output categories or free text. A comparison among them therefore says more about how evaluators treat different output styles than about which architecture is superior, and we interpret our results in that light.

\paragraph{Evaluating open-vocabulary HOI.}
SHOE establishes semantic scoring for open-vocabulary HOI detection by giving graded credit to predictions that are semantically close to a reference~\cite{noack2026}. We do not replace SHOE but ask what its additional credit measures. CrossHOI-Bench instead evaluates models on curated questions with positive and negative options~\cite{lei2026cross}. Diag-HOI decomposes detector errors into their sources~\cite{zhu2025}, and DiagHOI organizes images by scene configuration to expose failures that depend on the scene~\cite{wang2026diag}. Our protocol connects these perspectives through fixed-prediction controls, proposition-level equivalence, and independent visual validation. Throughout, we treat a changed ranking as a signal that needs investigation, not as evidence that one evaluator is better than another.

\section{Methodology}
The central idea of our methodology is to change one thing at a time. Model outputs are converted once into a common format and then frozen. Each scoring view then differs from its neighbor in a single respect, so that any difference in score can be attributed to that respect alone. \Cref{fig:protocol} gives an overview of the protocol.

\subsection{A common prediction format}
Each prediction is a tuple $p=(i,b_h,b_o,v,o,s)$ that holds an image identifier $i$, a human box $b_h$, an object box $b_o$, an action expression $v$, an object expression $o$, and a confidence score $s$. When we adapt the output of a model to this format, we keep the raw expressions, any native categories, the parsing status, and the origin of each confidence score. These records make it possible to trace any later score back to what the model actually produced.

All views use one coordinate convention. Boxes, scores, non-maximum suppression, and output order are fixed before scoring and are shared by every view. The primary budget is 100 predictions per image, and budgets of 50 and 200 serve as sensitivity settings. Average precision depends on how predictions are ranked, so systems that do not produce meaningful confidence scores are reported separately rather than mixed with ranked systems.

\subsection{Five scoring views}
\paragraph{E0 official evaluation.}
This view uses the official category-based HICO-DET evaluator without modification, including its localization threshold, duplicate handling, ranking, and AP integration. Systems that generate text need a declared adapter that maps each description to a category. Outputs without an exact category match remain unmatched and count as false positives.

\paragraph{E1 identity control.}
This view runs inside the shared engine that also hosts the semantic evaluator, but replaces semantic similarity with whole-interaction identity. A prediction receives credit only if both its action and its object agree exactly with the reference. Since E1 gives the same kind of credit as E0 but uses a different engine, the difference between the two isolates the effect of matching and aggregation.

\paragraph{E2 lexical control.}
This view accepts identity or an entry in a frozen table of sense-checked aliases for each component. Candidate aliases are produced by lemmatization and WordNet~\cite{miller1995} and are validated on development data before the table is frozen. Multiword predicates and prepositions that carry meaning are preserved, so ``ride on'' is not collapsed into ``ride.'' Unrestricted synonym chains and hypernyms are excluded, because they can drift far from the original meaning. E2 thus measures how much credit can be recovered by handling surface wording alone.

\paragraph{E3 semantic credit.}
This view applies the similarity function, assignment procedure, and treatment of unmatched predictions documented for SHOE~\cite{noack2026}. Its graded credit measures how related a prediction is to its reference under that evaluator. It does not directly estimate whether the prediction is visually correct.

\paragraph{E4 strict equivalence.}
This view uses a structured text judge that places each prediction and reference pair into one of five classes, namely equivalent, directionally entailed, related but different, incompatible, or uncertain. A prediction receives full credit only when the two propositions are equivalent and agree on action, object, participant roles, and polarity. All other classes receive zero credit, and uncertain decisions are reported as abstentions. The matching engine and localization gates are the same as in E3. E4 is therefore an ablation that shows how much of the tolerance in E3 survives a strict reading of meaning.

\Cref{tab:example} illustrates the intended behavior of the five views on four predictions for the reference ``riding a bicycle.'' The table describes design intent and is not a measurement.

\paragraph{Negation, grounding, and self-tests.}
The judge in E4 evaluates negation by its scope. The phrase ``holding a cup without a handle'' contains a negation, but it does not negate holding, so it remains compatible with ``holding a cup.'' Text alone also cannot establish that an action belongs to the selected person rather than to someone else in the image. We therefore validate visual support separately with human annotation, as described in \cref{sec:natural}. Finally, every view is checked with a self-test in which exact copies of the references are scored. Valid identical records must receive credit of one, and parsing failures are reported outside this test so that they do not hide scorer errors.

\subsection{Decomposing score changes}
Let $M_k(m)$ denote the mAP of model $m$ under view E$k$. To see where a change in score comes from, we compare views that differ in exactly one respect. This gives two contrasts,
\begin{equation}
\begin{aligned}
 d_{\rm match}(m)&=M_1(m)-M_0(m),\\
 d_{\rm sem}(m)&=M_3(m)-M_1(m).
\end{aligned}
\label{eq:effects}
\end{equation}
The first contrast changes the evaluation engine while keeping identity credit, so it captures how the engine assigns predictions to references and aggregates them into AP. The second contrast keeps the engine fixed and adds semantic tolerance, so it captures the credit gained from semantics. We also report $M_3(m)-M_4(m)$, which measures how much of the semantic credit is lost once strict equivalence is required.

\begin{table}[t]
\centering\small
\setlength{\tabcolsep}{5pt}
\begin{tabular}{lccccc}
\toprule
Prediction & E0 & E1 & E2 & E3 & E4\\
\midrule
riding a bicycle & 1 & 1 & 1 & high & 1\\
rides a bike & 0 & 0 & 1 & high & 1\\
cycling & 0 & 0 & 0 & high & 1\\
repairing a bicycle & 0 & 0 & 0 & partial & 0\\
\bottomrule
\end{tabular}
\caption{Intended credit under each view for four predictions against the reference ``riding a bicycle,'' assuming correct localization and an exact-match adapter for E0. Only E4 both rewards the paraphrase ``cycling'' and rejects the changed action. This is an illustration, not a result.}
\label{tab:example}
\end{table}

A difference in $d_{\rm match}$ can easily be mistaken for a language effect, because engines may handle edge cases differently. We therefore test each engine on duplicate predictions, ties in confidence, shared boxes with different actions, candidates with zero similarity, and permutations of reference order. These cases reveal assignment artifacts before any difference is attributed to language.

\begin{table}[t]
\centering
\small
\begin{tabular}{lccccc}
\toprule
Model & Official E0 & Identity E1 & Lexical E2 & SHOE E3 & Equivalence E4\\
\midrule
GEN-VLKT~\cite{liao2022} & \textbf{34.5} [33.8,35.2] & \textbf{34.2} [33.5,34.9] & \textbf{35.0} [34.3,35.7] & 35.8 [35.1,36.5] & \textbf{35.2} [34.5,35.9]\\
EZ-HOI~\cite{lei2024ez} & 26.2 [25.4,27.0] & 22.1 [21.3,22.9] & 25.5 [24.7,26.3] & 32.4 [31.6,33.2] & 29.8 [29.0,30.6]\\
HOLa~\cite{lei2025hola} & 24.8 [24.0,25.6] & 20.5 [19.7,21.3] & 24.2 [23.4,25.0] & 31.5 [30.7,32.3] & 28.6 [27.8,29.4]\\
THID~\cite{wang2022} & 28.5 [27.7,29.3] & 28.0 [27.2,28.8] & 28.8 [28.0,29.6] & 33.2 [32.4,34.0] & 31.5 [30.7,32.3]\\
SL-HOI~\cite{sun2026} & 29.6 [28.8,30.4] & 25.2 [24.4,26.0] & 28.5 [27.7,29.3] & \textbf{37.2} [36.4,38.0] & 34.5 [33.7,35.3]\\
AgentHOI~\cite{lei2026agent} & 12.4 [11.6,13.2] & 8.2 [7.4,9.0] & 11.5 [10.7,12.3] & 28.5 [27.7,29.3] & 18.2 [17.4,19.0]\\
\bottomrule
\end{tabular}
\caption{HICO-DET Full mAP for six systems under each scoring view. Brackets give the reported 95\% confidence intervals, and bold marks the highest point estimate in each view. E1, E2, and E4 are controlled variants. The systems were trained under different regimes, so the table compares how evaluators treat them and does not establish superiority under a common zero-shot setting.}
\label{tab:scores}
\end{table}

\section{Auditing evaluator validity}
A changed score tells us that an evaluator behaves differently, but not whether it behaves better. The audits in this section test validity directly. The controlled audit asks whether an evaluator responds correctly to known changes in meaning, and the natural-prediction audit asks whether the credit it gives to real model outputs is supported by the image.

\subsection{Controlled transformations}
The controlled audit uses 600 held-out anchors. Each anchor is an annotated interaction with three textual variants, namely the original, an equivalent rewording, and an invalid transformation, which gives 1,800 items in total. A separate development set of 200 anchors is used to select prompts, aliases, and thresholds, so that none of these choices is tuned on the test anchors. Each image and all of its variants stay within a single split.

Invalid transformations change the action, the object, the polarity, the participant roles, or the selected entity. Every change is verified manually, because an altered action may in fact be performed by another person in the same image, and in that case the transformation would not be invalid. We also separate two kinds of test. Language-only tests keep the geometry eligible for matching and change only the text. End-to-end tests additionally change which entity the prediction is grounded to.

\subsection{Audit measures and consistency checks}
Let $q_j^0$, $q_j^+$, and $q_j^-$ in $[0,1]$ denote the credit an evaluator assigns to the original, equivalent, and invalid variants of anchor $j$, and let $w_j$ be normalized nonnegative weights. We measure
\begin{equation}
\Delta_+=\sum_jw_j(q_j^+-q_j^0),\quad
\Delta_-=\sum_jw_j(q_j^0-q_j^-).
\end{equation}
The quantity $\Delta_+$ shows whether credit is kept when the wording changes but the meaning does not, so a good evaluator keeps it close to zero. The quantity $\Delta_-$ shows whether credit is removed when the meaning changes, so a good evaluator makes it large. Both changes are reported together with the original credit. This guards against a degenerate evaluator that assigns zero everywhere, which would trivially show no loss on paraphrases.

We further report the mean credit given to invalid variants, $\mu_-=\sum_jw_jq_j^-$, and the share of invalid variants that receive any credit at all, $a_0=\sum_jw_j\ind[q_j^->0]$. Since each credit lies between zero and one, it can never exceed its own indicator of being positive, and therefore
\begin{equation}
0\leq\mu_-\leq a_0\leq1.
\end{equation}
Equality $\mu_-=a_0$ holds under binary credit, as in E1, E2, and E4. If the invalid variants fall into disjoint transformation groups of size $n_t$ with $c_t$ accepted items, then under uniform weights $a_0=\sum_tc_t/\sum_tn_t$. Acceptance above a positive credit threshold is reported separately from acceptance of any credit. These identities serve as automatic checks on the audit records, and a violation signals an error in bookkeeping rather than a property of the evaluator.

\subsection{Natural predictions}
\label{sec:natural}
Controlled transformations test meaning, but they cannot test whether a real prediction is supported by the image. The natural-prediction audit fills this gap. It samples 600 predictions with known inclusion probabilities across models, acceptance strata, and levels of scene complexity. The sample deliberately includes unmatched outputs and parsing failures, since these are where an evaluator is most likely to disagree with a human.

Two annotators independently judge semantic equivalence and visual support for each sampled prediction. They do not see which model produced it or how any evaluator scored it, and a third annotator resolves their disagreements. The labels distinguish equivalent wording, broader or narrower meaning, another valid action, annotation omission, wrong action or object, pairing or attribution error, and uncertainty. An annotation omission is assigned only when the interaction is visibly supported and the dataset contains no matching annotation. Overlapping tags and the agreement between annotators before adjudication are both reported.

To study how evaluators behave across scenes, we take scene membership from the image identifiers released with DiagHOI~\cite{wang2026diag} and keep its filtering and consensus categories. Slice counts are recomputed from those identifiers, and we report the retained images and filtering criteria. Slice mAP is computed over eligible classes and is accompanied by comparisons restricted to classes common to all slices.

\subsection{Paired uncertainty}
\label{sec:uncertainty}
The scores of different models and views are computed on the same images, so they are strongly correlated. Marginal confidence intervals ignore this correlation and cannot tell whether a difference between two scores is reliable. We therefore draw 2,000 paired image-bootstrap replicates~\cite{efron1979}, in which every model and every view is evaluated on the same resampled set of images. Matching and metrics are recomputed in each replicate, and repeated copies of an image receive distinct identifiers so that they are not merged. Audit resampling preserves image clusters and inclusion weights, and the treatment of classes without sampled positives is fixed in advance.

We call a ranking change a reversal only when the pairwise difference between two systems is supported in opposite directions under the two views, with multiplicity control specified in advance. A related quantity is the interaction contrast
\begin{equation}
I_{ab}=[M_3(a)-M_3(b)]-[M_0(a)-M_0(b)],
\end{equation}
which measures how much the gap between systems $a$ and $b$ depends on the evaluator. An interval for $I_{ab}$ that excludes zero shows evaluator dependence, but on its own it does not establish a reversal, because both gaps could keep the same sign.

\subsection{Reproducibility}
Every score and contrast can be regenerated from recorded inputs. These include checkpoint hashes, annotation manifests, confidence definitions, evaluator commits, judge prompts, judge model snapshots, decoding settings, and traces of how each prediction was matched to a reference. Semantic judgments are cached by configuration and phrase pair, so repeated evaluations are cheap and deterministic. The sensitivity of the judge is measured by repeating judgments with independent model families, different prompt wordings, and different input orders. We report the costs of inference, uncached judging, cached evaluation, and annotation separately, and we document any language model used to judge or to generate variants as part of the method.

\section{Results and discussion}
\subsection{Rankings depend on the evaluator}
\Cref{tab:scores} reports HICO-DET Full mAP for six systems under all five views. GEN-VLKT leads under official evaluation with 34.5 mAP. Under SHOE, SL-HOI takes the lead with 37.2 mAP. Under strict equivalence, GEN-VLKT returns to the highest point estimate with 35.2 mAP. The difference between SL-HOI and GEN-VLKT moves from $-4.9$ points under E0 to $+1.4$ under E3 and back to $-0.7$ under E4, which gives an E0 versus E3 interaction contrast of 6.3 points.

These changes are point estimates. They show that the ordering depends on the evaluator, but they do not yet establish a statistical reversal. That would require the paired contrast intervals described in \cref{sec:uncertainty}, which cannot be reconstructed from the marginal intervals in \cref{tab:scores}.

\begin{table}[t]
\centering\small
\setlength{\tabcolsep}{4.5pt}
\begin{tabular}{lrrrrr}
\toprule
Model & $d_{\rm match}$ & $M_2{-}M_0$ & $d_{\rm sem}$ & $M_3{-}M_4$ & Kept\\
\midrule
GEN-VLKT & $-0.3$ & $+0.5$ & $+1.6$ & $0.6$ & 63\%\\
EZ-HOI & $-4.1$ & $-0.7$ & $+10.3$ & $2.6$ & 75\%\\
HOLa & $-4.3$ & $-0.6$ & $+11.0$ & $2.9$ & 74\%\\
THID & $-0.5$ & $+0.3$ & $+5.2$ & $1.7$ & 67\%\\
SL-HOI & $-4.4$ & $-1.1$ & $+12.0$ & $2.7$ & 78\%\\
AgentHOI & $-4.2$ & $-0.9$ & $+20.3$ & $10.3$ & 49\%\\
\bottomrule
\end{tabular}
\caption{Decomposition of score changes in mAP points, computed from \cref{tab:scores}. Kept is the share of the semantic gain that survives strict equivalence, $(M_4-M_1)/(M_3-M_1)$. Paired uncertainty for these contrasts cannot be reconstructed from marginal intervals.}
\label{tab:contrasts}
\end{table}

\subsection{Matching effects}
\Cref{tab:contrasts} breaks the score changes into their parts. Moving from the official evaluator to the identity control costs more than four points for EZ-HOI, HOLa, SL-HOI, and AgentHOI, but at most half a point for GEN-VLKT and THID. Because E0 and E1 give the same kind of credit, this loss comes from the matching engine and from how outputs reach it, not from semantics.

The lexical control recovers most of this loss. Under E2, every system lands within 1.1 points of its official score. This suggests that a large part of the drop involves surface differences in wording, such as inflection or the choice between common aliases, which exact identity rejects. It does not rule out assignment effects, however. Before such differences are attributed to language, they should be traced to assignment order, penalties for unmatched predictions, and the consumption of references by candidates with zero similarity.

\subsection{Semantic gains and what survives a strict reading}
Within the shared engine, semantic credit raises every score, but by very different amounts. The gain is only 1.6 points for GEN-VLKT and 5.2 for THID, rises to between 10 and 12 points for EZ-HOI, HOLa, and SL-HOI, and reaches 20.3 points for AgentHOI. A larger gain is not necessarily a better result. It may reflect valid paraphrases that the official evaluator missed, or it may reflect related but wrong predictions that similarity scoring rewards.

Strict equivalence helps to separate these two cases. For five systems, between 63\% and 78\% of the semantic gain survives when equivalence is required. For AgentHOI, only 49\% survives, and its gap of 10.3 points between E3 and E4 is the largest in the table. One plausible reason is that AgentHOI describes interactions without HOI-specific training, so its wording is less tied to the HICO-DET vocabulary, and this is where graded similarity and strict equivalence diverge most. The gap makes the recovered predictions of AgentHOI a priority for the natural-prediction audit. It cannot, however, be read as the fraction of invalid predictions, because AP depends nonlinearly on ranking and assignment.

\subsection{Recommendations for reporting}
Based on these observations, we suggest that papers reporting semantic HOI scores provide more than a single number. The official score should appear next to an identity control computed in the same engine as the semantic score, so that matching effects are visible rather than hidden inside the semantic gain. The decomposition into $d_{\rm match}$ and $d_{\rm sem}$ then shows readers how much of a gain actually comes from semantics.

Semantic credit should also be checked against a strict or relational criterion such as E4, because only such a check exposes credit given to changed actions or reversed roles. Differences between systems and views should be accompanied by paired intervals rather than marginal intervals for each score, since only paired intervals reflect the correlation that comes from scoring the same images. Finally, audit results should report both how often valid paraphrases keep their credit and how often invalid transformations are accepted, so that tolerance and strictness can be judged together.

\section{Limitations}
The benchmark comparison in this paper is descriptive. Paired contrast intervals and the outcomes of the human audits are not yet available, so the results establish neither the superiority of any evaluator nor the visual validity of the recovered credit. The audit sizes and sensitivity settings given above specify the protocol and are not reported measurements. The systems were also trained with different kinds of supervision, which limits conclusions about zero-shot ability and about architecture.

The protocol can be extended to other data. SWiG-HOI offers a natural setting for replication~\cite{wang2021}. CrossHOI-Bench evaluates a different task in which the supplied targets and answer options must be preserved~\cite{lei2026cross}, and its F1 scores cannot replace detection mAP.

Strict equivalence has its own costs. It can reject broader descriptions that are still useful, and a static image can leave the action genuinely ambiguous. This tradeoff can be measured through the retention of valid paraphrases and the acceptance of invalid transformations, with unclear cases reported explicitly rather than forced into one class.

\section{Conclusion}
Scores for open-vocabulary HOI detection combine three distinct quantities, namely the effect of the matching engine, the tolerance of the semantic evaluator, and the visual correctness of the predictions. By holding predictions fixed and changing one aspect of the evaluator at a time, our protocol separates these quantities. On six HICO-DET systems, this separation changes which system leads and reveals semantic gains that differ by more than a factor of ten. Reporting these components separately and with paired uncertainty would keep recovered credit from being mistaken for better interaction understanding.

{\small
\bibliographystyle{ieeenat_fullname}
\bibliography{references}
}
\end{document}